\documentclass[10pt,twocolumn,letterpaper]{article}

\usepackage[pagenumbers]{cvpr} %

\definecolor{cvprblue}{rgb}{0.21,0.49,0.74}
\usepackage[pagebackref,breaklinks,colorlinks,allcolors=cvprblue]{hyperref}

\newcommand{\modelname}{1.58-bit FLUX\xspace}

\def\paperID{*****} %
\def\confName{CVPR}
\def\confYear{2025}

\title{\modelname
}

\author{\centerline{Chenglin Yang\textsuperscript{1}, Celong Liu\textsuperscript{1}, Xueqing Deng\textsuperscript{1}, Dongwon Kim\textsuperscript{2},} \\
\centerline{Xing Mei\textsuperscript{1}, Xiaohui Shen\textsuperscript{1}, Liang-Chieh Chen\textsuperscript{1}}\\
\\
\textsuperscript{1}ByteDance\quad\textsuperscript{2}POSTECH\\
\url{https://chenglin-yang.github.io/1.58bit.flux.github.io/}
}

\usepackage{amsmath}
\usepackage{bbm}
\usepackage{multirow}
\usepackage{xcolor}

\newcommand{\figref}[1]{Fig.~\ref{#1}}
\newcommand{\tabref}[1]{Tab.~\ref{#1}}

\usepackage{colortbl}

\newcolumntype{L}[1]{>{\raggedright\let\newline\\\arraybackslash\hspace{0pt}}m{#1}}
\newcolumntype{C}[1]{>{\centering\let\newline\\\arraybackslash\hspace{0pt}}m{#1}}
\newcolumntype{R}[1]{>{\raggedleft\let\newline\\\arraybackslash\hspace{0pt}}m{#1}}

\begin{document}

\twocolumn[{
\renewcommand\twocolumn[1][]{#1}
\maketitle
\begin{center}
 \centering
 \small
 \setlength{\tabcolsep}{0.0pt}
 \begin{tabular}{c}
    \includegraphics[width=1.0\textwidth]{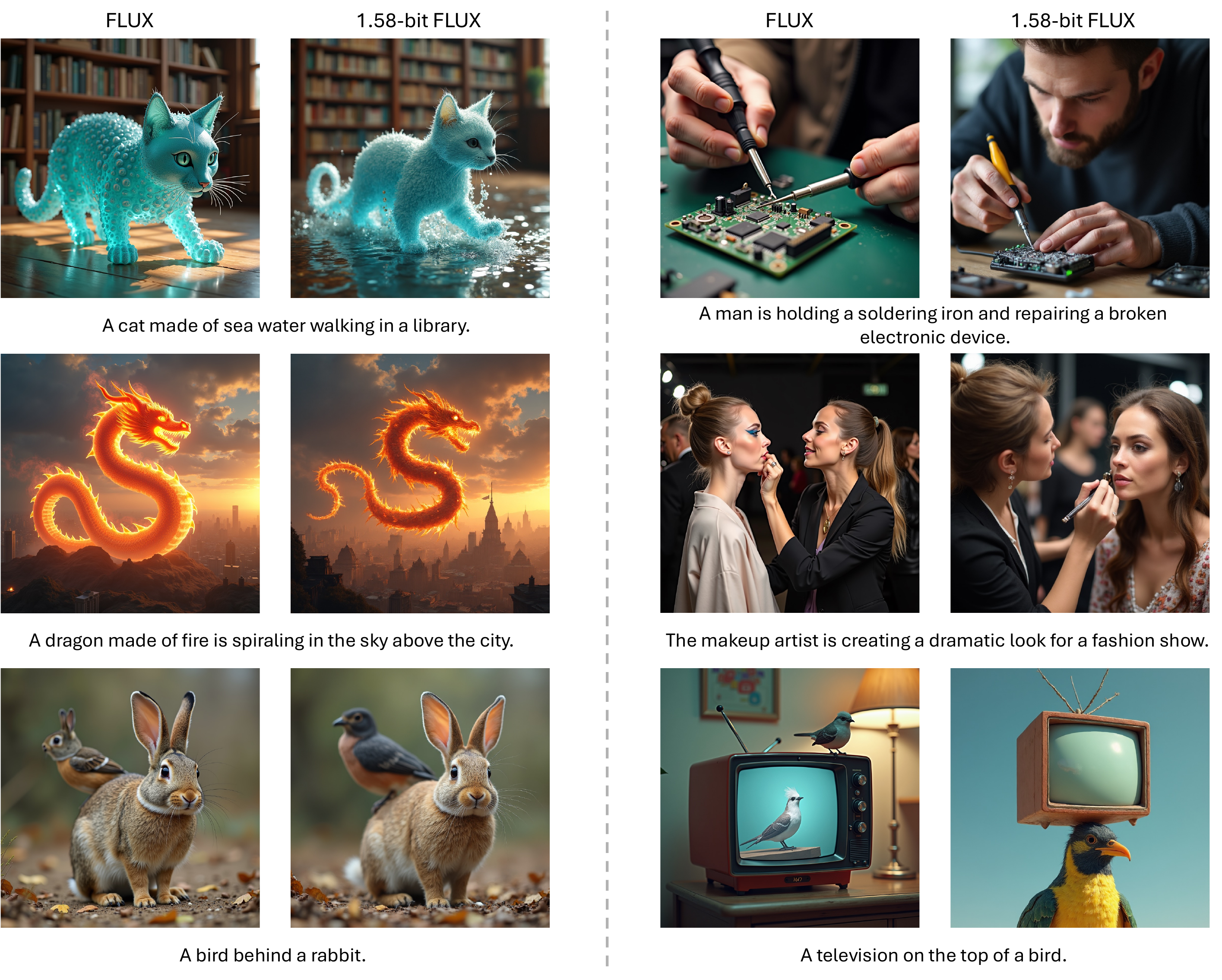}
 \end{tabular}
 \captionof{figure}{
 \textbf{Visual comparisons between FLUX and \modelname.}
 \modelname demonstrates comparable generation quality to FLUX while employing 1.58-bit quantization, where 99.5\% of the 11.9B parameters in the vision transformer are constrained to the values +1, -1, or 0. For consistency, all images in each comparison are generated using the same latent noise input. \modelname utilizes a custom 1.58-bit kernel. Additional visual comparisons are provided in~\figref{fig: vis_compare_geneval} and~\figref{fig: vis_compare_t2i}.
 }
 \label{fig: vis_compare_1}
\end{center}
}]

\twocolumn[{
\renewcommand\twocolumn[1][]{#1}
\maketitle
\begin{center}
\centering
 \small
 \setlength{\tabcolsep}{0.0pt}
 \begin{tabular}{c}
    \includegraphics[width=0.85\textwidth]{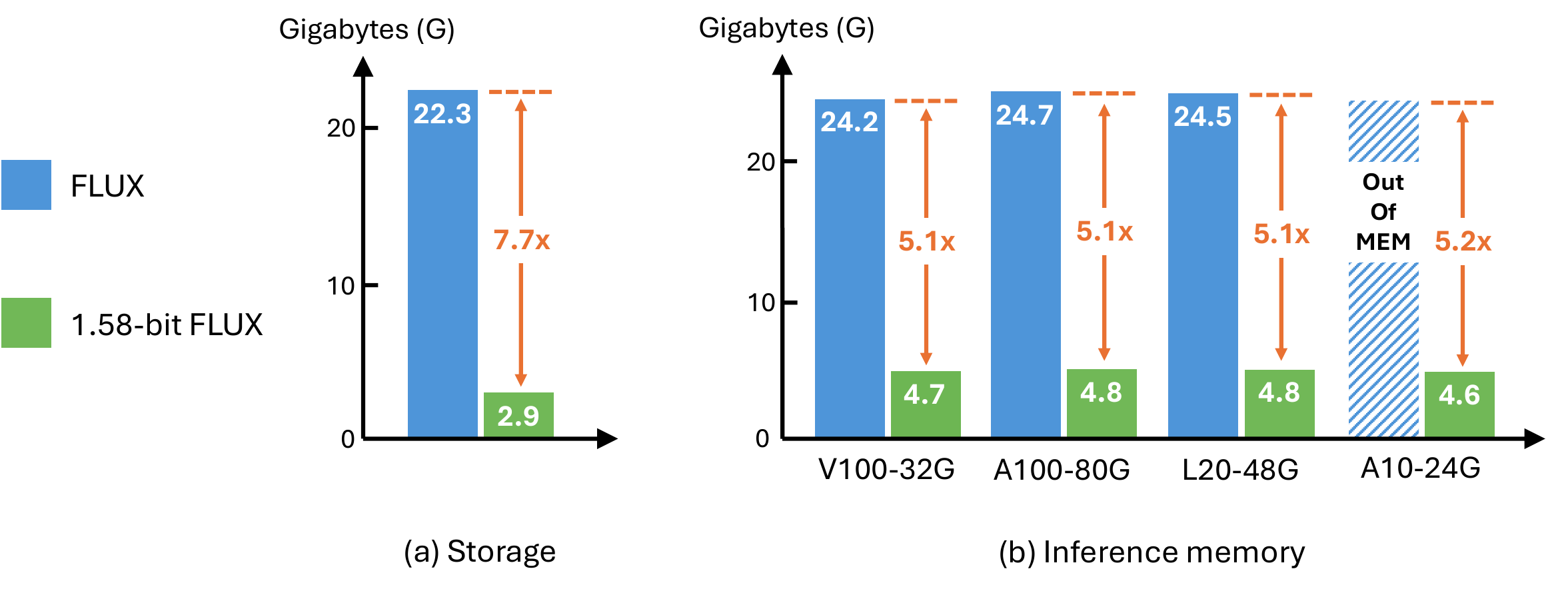}
 \end{tabular}
 \captionof{figure}{
 \textbf{Efficiency measurements on the vision transformer component of FLUX and \modelname.}
 The measurements are based on generating a single image with 50 inference steps.
(a) \modelname reduces checkpoint storage by 7.7× compared to FLUX.
(b) \modelname achieves a 5.1× reduction in inference memory usage across various GPU types.
The x-axis labels, $m$-$n$G, represent GPU type $m$ with a maximum memory capacity of $n$ Gigabytes (G).
 }
 \label{fig: efficiency_compare}
\end{center}
}]

\maketitle
\begin{abstract}

We present \modelname, the first successful approach to quantizing the state-of-the-art text-to-image generation model, FLUX.1-dev, using 1.58-bit weights (\ie, values in \{-1, 0, +1\}) while maintaining comparable performance for generating 1024 × 1024 images. Notably, our quantization method operates without access to image data, relying solely on self-supervision from the FLUX.1-dev model. Additionally, we develop a custom kernel optimized for 1.58-bit operations, achieving a 7.7× reduction in model storage, a 5.1× reduction in inference memory, and improved inference latency. Extensive evaluations on the GenEval and T2I Compbench benchmarks demonstrate the effectiveness of \modelname in maintaining generation quality while significantly enhancing computational efficiency.
\end{abstract}

\begin{table*}[t]
\centering
\resizebox{1.0\textwidth}{!}{%
\begin{tabular}{@{}cc|cccccccc@{}}
    \toprule 
    \multirow{3}{*}{Model} & \multirow{3}{*}{Avg.}  & \multirow{2}{*}{Color} & \multirow{2}{*}{Shape} & \multirow{2}{*}{Texture} & 2D & 3D & \multirow{2}{*}{Numeracy} & Non- & \multirow{2}{*}{Complex} \\
    {} & {} & {} & {} & {} & Spatial & Spatial & {} & spatial & {} \\
    \cmidrule(lr){3-3} \cmidrule(lr){4-4} \cmidrule(lr){5-5} \cmidrule(lr){6-6} \cmidrule(lr){7-7} \cmidrule(lr){8-8} \cmidrule(lr){9-9} \cmidrule(lr){10-10} 
    {} & {} & B-VQA & B-VQA & B-VQA & UniDet & UniDet & UniDet & S-CoT & S-CoT \\
    \midrule
    Stable XL~\cite{podell2023sdxl} & 0.5255 & 0.5879 & 0.4687 & 0.5299 & 0.2133 & 0.3566 & 0.4988 & 0.7673 & 0.7817 \\
    Pixart-$\alpha$-ft~\cite{chen2023pixart} & 0.5586 & 0.6690 & 0.4927 & 0.6477 & 0.2064 & 0.3901 & 0.5058 & 0.7747 & 0.7823  \\
    \midrule
    FLUX & 0.5876 & 0.7529 & 0.5056 & 0.6299 & 0.2791 & 0.4014 & 0.6131 & 0.7807 & 0.7380  \\
    \modelname (w/o kernel) & 0.5806 & 0.7358 & 0.4900 & 0.6151 & 0.2781 & 0.4037 & 0.6089 & 0.7807 & 0.7327  \\
    \modelname & 0.5812 & 0.7390 & 0.4910 & 0.6162 & 0.2757 & 0.4049 & 0.6137 & 0.7793 & 0.7300 \\
    \bottomrule
\end{tabular}
}
\caption{\textbf{Evaluations on T2I CompBench.} \modelname (w/o kernel) indicates no efficient kernel is applied.}
\label{tab: t2i_compbench}
\end{table*}

\begin{table*}[t]
\centering
\begin{tabular}{@{}cc|cccccc@{}}
    \toprule
    \multirow{2}{*}{Model} & \multirow{2}{*}{Overall} & Single & Two & \multirow{2}{*}{Counting} & \multirow{2}{*}{Colors} & \multirow{2}{*}{Position} & Color \\
    {} & {} & object & object & {} & {} & {} & attribution \\
    \midrule
    Stable XL~\citep{podell2023sdxl}         & 0.55   & 0.98   & 0.74   & 0.39   & 0.85  & 0.15 & 0.23 \\
    PlayGroundv2.5~\citep{li2024playground} & 0.56   & 0.98   & 0.77   & 0.52   & 0.84  & 0.11 & 0.17 \\
    \midrule
    FLUX & 0.66 & 0.98 & 0.81 & 0.74 & 0.79 & 0.22 & 0.45 \\
    \modelname (w/o kernel) & 0.64 & 0.98 & 0.79 & 0.69 & 0.77 & 0.20 & 0.43  \\
    \modelname & 0.64 & 0.98 & 0.77 & 0.68 & 0.79 & 0.21 & 0.44  \\
    \bottomrule
\end{tabular}
\caption{\textbf{Evaluations on GenEval.} \modelname (w/o kernel) indicates no efficient kernel is applied.}
\label{tab: geneval_compare}
\end{table*}

\begin{table}[t]
\centering
\begin{tabular}{@{}l|ccc@{}}
    \toprule 
    GPU & FLUX & \modelname & Improvements \\
    \midrule
    V100 & 74.8 & 73.6 & 1.6\% \\
    A100 & 26.4 & 25.0 & 5.3\% \\
    L20 & 90.2 & 78.3 & 13.2\% \\
    A10 & OOM & 84.4 & -- \\
    \bottomrule
\end{tabular}
\caption{\textbf{Latency measurements on the vision transformer component of FLUX and \modelname.} The measurements are obtained by generating one image with 50 inference steps. OOM means out of memory.}
\label{tab: latency_measurements}
\end{table}

\section{Introduction}
\label{sec:intro}

Recent text-to-image (T2I) models, including DALLE 3~\cite{betker2023improving}, Adobe Firefly 3~\cite{adobe2024firefly3}, Stable Diffusion 3~\cite{esser2024scaling}, Midjourney v6.1~\cite{midjourney2024v6.1}, Ideogram v2~\cite{ideogram2024v2.0}, PlayGround V3~\cite{liu2024playground}, FLUX 1.1~\cite{black2024flux1.1}, Recraft V3~\cite{recraft2024}, have demonstrated remarkable generative capabilities, making them highly promising for real-world applications. However, their immense parameter counts, often in the billions, and high memory requirements during inference pose significant challenges for deployment on resource-constrained devices such as mobile platforms.

In this work, we address these challenges by exploring extreme low-bit quantization of T2I models. Among available state-of-the-art models, we select FLUX.1-dev~\cite{black2024fluxdev} as our quantization target due to its public availability and competitive performance.\footnote{We refer FLUX to FLUX.1-dev in this paper for simplicity.} Specifically, we quantize the weights of the vision transformer in FLUX to 1.58 bits without relying on mixed-precision schemes or access to image data. The quantization restricts linear layer weights to the values \{+1, 0, -1\}, akin to the BitNet b1.58~\cite{ma2024era} approach. Unlike BitNet, which involves training large language models from scratch, our method operates as a post-training quantization solution for T2I models.

This quantization reduces model storage by 7.7$\times$, as the 1.58-bit weights are stored using 2-bit signed integers, compressing them from 16-bit precision. To further enhance inference efficiency, we introduce a custom kernel optimized for low-bit computation. This kernel reduces inference memory usage by over 5.1$\times$ and improves inference latency, as detailed in~\figref{fig: efficiency_compare} and~\tabref{tab: latency_measurements}. Comprehensive evaluations on T2I benchmarks, including GenEval~\cite{ghosh2024geneval} and T2I Compbench~\cite{huang2023t2i}, reveal that \modelname achieves comparable performance to full-precision FLUX, as shown in~\tabref{tab: t2i_compbench} and~\tabref{tab: geneval_compare}.

Our contributions can be summarized as follows:

\begin{itemize}[]
\item We introduce \modelname, the first quantized model to reduce 99.5\% of FLUX vision transformer parameters (11.9B in total) to 1.58 bits without requiring image data, significantly lowering storage requirements.
\item We develop an efficient linear kernel optimized for 1.58-bit computation, enabling substantial memory reduction and inference speedup.
\item We demonstrate that \modelname maintains performance comparable to the full-precision FLUX model on challenging T2I benchmarks.
\end{itemize}
This work represents a significant step forward in making high-quality T2I models practical for deployment on memory- and latency-constrained devices.

\begin{table*}
 \centering
 \small
 \setlength{\tabcolsep}{0.0pt}
 \begin{tabular}{c}
    \includegraphics[width=0.97\textwidth]{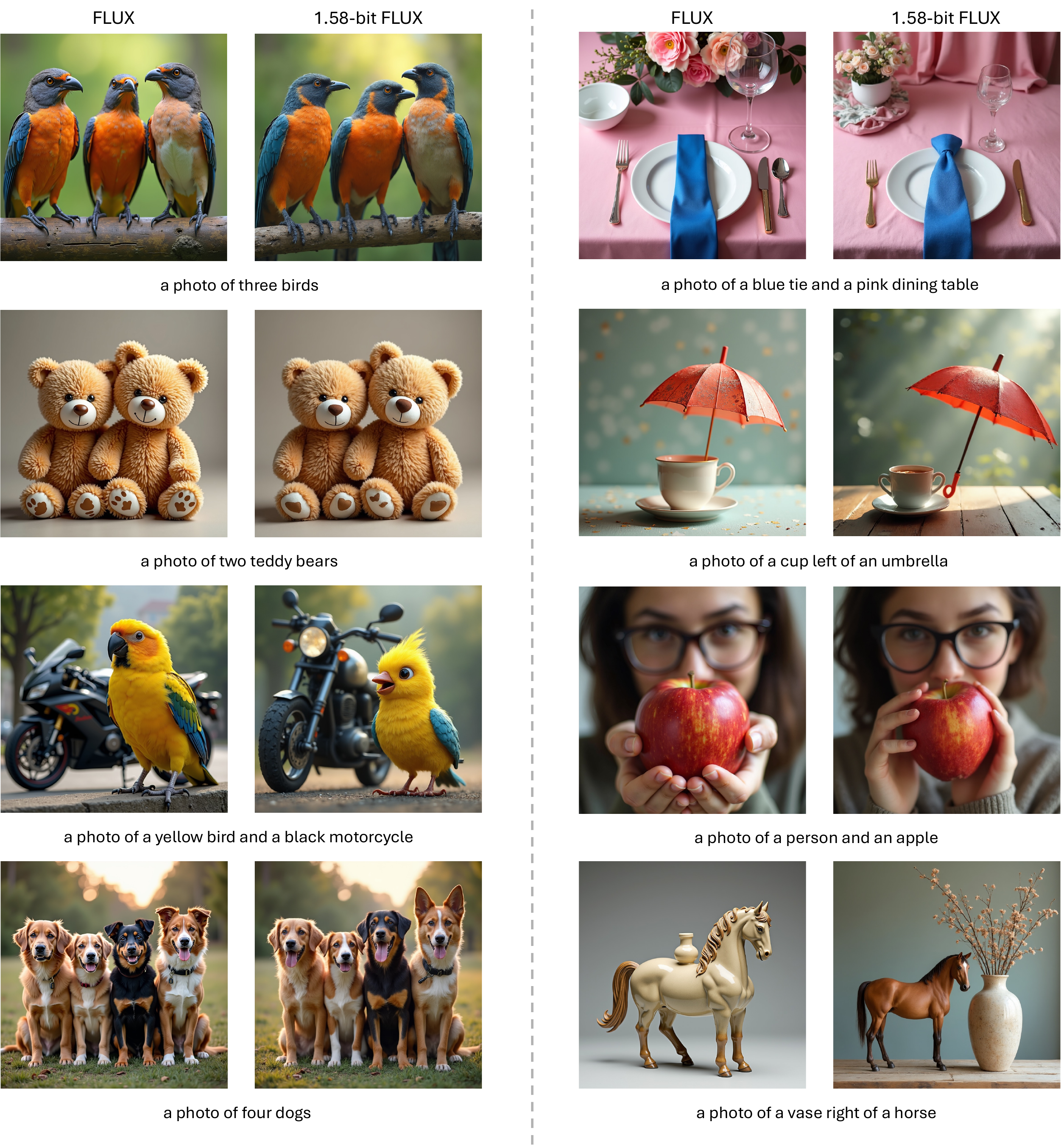}
 \end{tabular}
 \captionof{figure}{
 \textbf{Visual comparisons between FLUX and \modelname on GenEval dataset.}
 \modelname demonstrates comparable generation quality to FLUX while employing 1.58-bit quantization, where 99.5\% of the 11.9B parameters in the vision transformer are constrained to the values +1, -1, or 0. For consistency, all images in each comparison are generated using the same latent noise input. \modelname utilizes a custom 1.58-bit kernel.
 }
 \label{fig: vis_compare_geneval}
\end{table*}

\begin{table*}
 \centering
 \small
 \setlength{\tabcolsep}{0.0pt}
 \begin{tabular}{c}
    \includegraphics[width=0.97\textwidth]{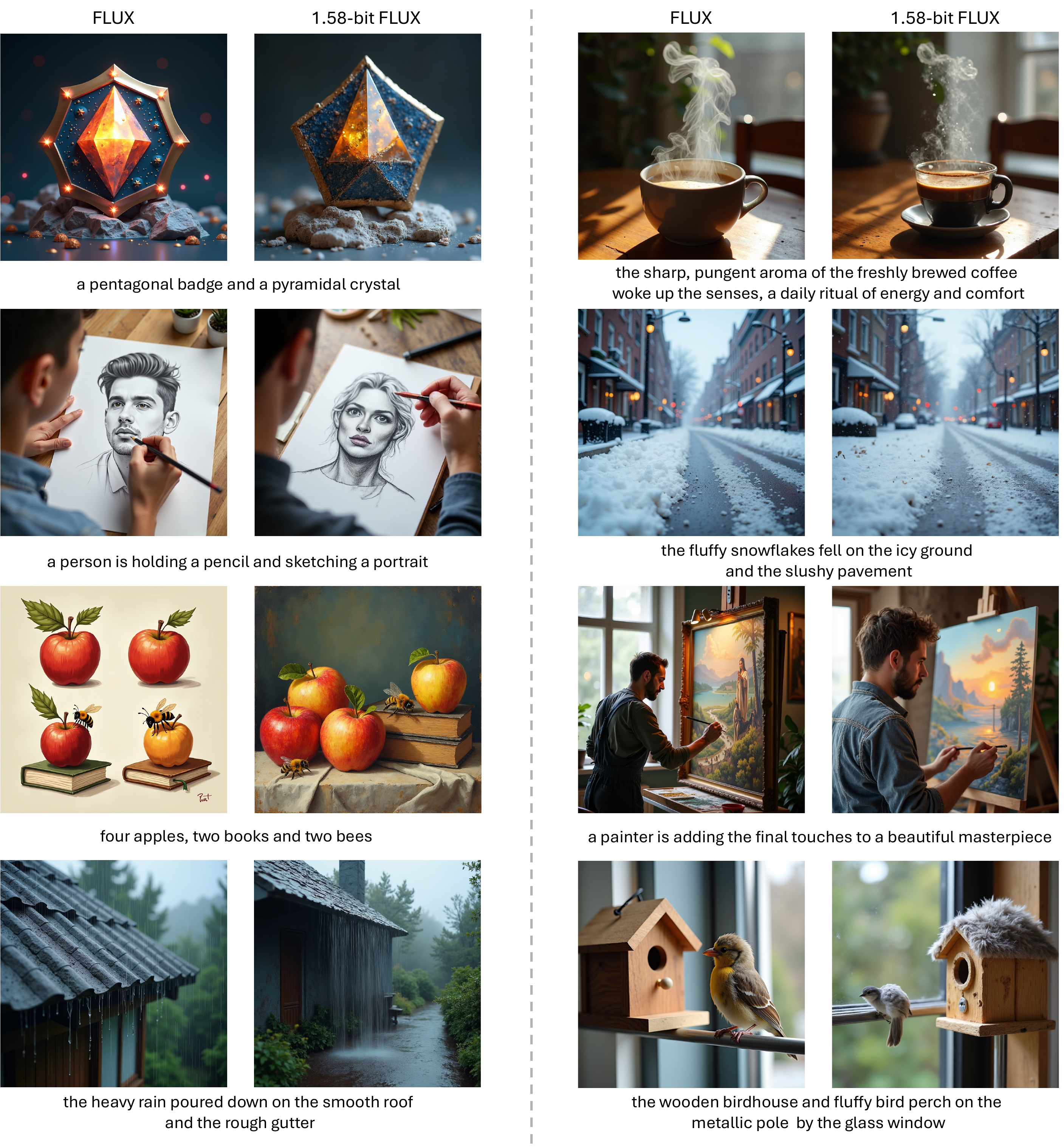}
 \end{tabular}
 \captionof{figure}{
 \textbf{Visual comparisons between FLUX and \modelname on the validation split of T2I CompBench.}
 \modelname demonstrates comparable generation quality to FLUX while employing 1.58-bit quantization, where 99.5\% of the 11.9B parameters in the vision transformer are constrained to the values +1, -1, or 0. For consistency, all images in each comparison are generated using the same latent noise input. \modelname utilizes a custom 1.58-bit kernel.
 }
 \label{fig: vis_compare_t2i}
\end{table*}

\section{Related Work}
\label{sec:related_work}

Quantization is a widely adopted technique for reducing model size and enhancing inference efficiency, as demonstrated in numerous studies~\cite{jacob2018quantization,hubara2016binarized,gong2019differentiable,nagel2021white,esser2019learned,nagel2020up,dong2019hawq,wang2019haq,han2015deep,choukroun2019low,cai2020zeroq}.
It has proven particularly effective for serving large language models (LLMs)~\cite{dettmers2022gpt3,xiao2023smoothquant,liu2023llm,yao2022zeroquant,frantar2022gptq,lin2024awq,wei2022outlier,shao2023omniquant,zhao2024atom,lin2024qserve,kim2023squeezellm,chen2024prefixquant}, enabling significant resource savings without compromising performance. Recent advancements include both post-training quantization (PTQ) methods~\cite{frantar2022gptq,xiao2023smoothquant,yao2022zeroquant}, which adjust pre-trained models for efficient deployment, and Quantization Aware Training (QAT) approaches that fine-tune the model to low bits from pretrained
checkpoints~\cite{liu2023llm,nagel2022overcoming,he2023bivit,chen2024efficientqat} 
or train the model from scratch~\cite{wang2023bitnet}. For instance, BitNet b1.58~\cite{ma2024era} employs weights in linear layers restricted to three values \{+1, 0, -1\}, facilitating highly efficient inference, particularly on CPUs~\cite{wang20241}. These developments highlight the potential of quantization to address the computational challenges associated with deploying large-scale models.

For image generation models, prior work has explored various quantization techniques for diffusion models~\cite{chen2024low,li2023q,shang2023post,he2024ptqd,yang2023efficient,huang2024tfmq,wang2024towards,liu2024enhanced,tang2025post,wu2024ptq4dit,he2023efficientdm,yao2024timestep,chen2024q}, including Hadamard transformations~\cite{liu2024hq}, vector quantization~\cite{deng2024vq4dit}, floating-point quantization~\cite{liu2024hq}, mixed bit-width allocation~\cite{zhao2024mixdq}, diverse quantization metrics~\cite{zhao2024vidit}, multiple-stage fine-tuning~\cite{sui2024bitsfusion} and low-rank branch~\cite{li2024svdquant}. These approaches aim to optimize model efficiency while maintaining performance. In this work, we focus on post-training 1.58-bit quantization of FLUX, a state-of-the-art open-source text-to-image (T2I) model. Notably, our method achieves efficient quantization without relying on any tuning image data and is complemented by optimized inference techniques.

\section{Experimental Results}
\label{sec:exp}

\subsection{Settings}

\noindent\textbf{Quantization.}
We utilize a calibration dataset comprising prompts from the Parti-1k dataset~\cite{yu2022scaling} and the training split of T2I CompBench~\cite{huang2023t2i}, totaling 7,232 prompts. This process is entirely image-data-free, requiring no additional datasets. The quantization reduces the weights of all linear layers in the FluxTransformerBlock and FluxSingleTransformerBlock of FLUX to 1.58 bits, covering 99.5\% of the model’s total parameters.

\noindent\textbf{Evaluation.}
We evaluate both FLUX and \modelname on the GenEval dataset~\cite{ghosh2024geneval} and the validation split of T2I CompBench~\cite{huang2023t2i}, following the official image generation pipeline. The GenEval dataset consists of 553 prompts, with four images generated per prompt. The T2I CompBench validation split includes eight categories, each containing 300 prompts, with 10 images generated per prompt, yielding a total of 24,000 images for evaluation. All images are generated at a resolution of 1024 × 1024 for both FLUX and \modelname.

\subsection{Results}

\noindent\textbf{Performance.} 
Comparable performances between \modelname and FLUX on T2I Compbench and GenEval are observed in~\tabref{tab: t2i_compbench} and~\tabref{tab: geneval_compare}, respectively. The minor differences observed before and after applying our linear kernel further demonstrate the accuracy of our implementation.

\noindent\textbf{Efficiency.} Significant efficiency gains are observed in both model storage and inference memory, as shown in~\figref{fig: efficiency_compare}. For inference latency, as illustrated in~\tabref{tab: latency_measurements}, even greater improvements are achieved when running \modelname on lower-performing yet deployment-friendly GPUs, such as the L20 and A10.

\section{Conclusion and Discussion}
\label{sec:conclusion}

This work introduced \modelname, in which 99.5\% of the transformer parameters are quantized to 1.58 bits. With our custom computation kernels, \modelname achieves a 7.7$\times$ reduction in model storage and more than a 5.1$\times$ reduction in inference memory usage. Despite these compression gains, \modelname demonstrates comparable performance on T2I benchmarks and maintains high visual quality. We hope that \modelname inspires the community to develop more robust models for mobile devices.

We discuss the current limitations of \modelname below, which we plan to address in future work:

\begin{itemize}
    \item \textbf{Limitations on speed improvements.} Although \modelname reduces model size and memory consumption, its latency is not significantly improved due to the absence of activation quantization and lack of further optimized kernel implementations. 
    Given our promising results, we hope to inspire the community to develop custom kernel implementation for 1.58-bit models.
    \item \textbf{Limitations on visual qualities.} \figref{fig: vis_compare_1},~\figref{fig: vis_compare_geneval} and~\figref{fig: vis_compare_t2i} show that \modelname can generate vivid and realistic images closely aligned with the given text prompts. However, it still lags behind the original FLUX model in rendering fine details at very high resolutions. We aim to address this gap in future research.
\end{itemize}

{
    \small
    \bibliographystyle{ieeenat_fullname}
    \bibliography{main}
}

\end{document}